\documentclass[letterpaper]{article} % DO NOT CHANGE THIS
\usepackage{aaai2026}  % DO NOT CHANGE THIS
\usepackage{times}  % DO NOT CHANGE THIS
\usepackage{helvet}  % DO NOT CHANGE THIS
\usepackage{courier}  % DO NOT CHANGE THIS
\usepackage[hyphens]{url}  % DO NOT CHANGE THIS
\usepackage{graphicx} % DO NOT CHANGE THIS
\usepackage{natbib}  % DO NOT CHANGE THIS AND DO NOT ADD ANY OPTIONS TO IT
\usepackage{caption} % DO NOT CHANGE THIS AND DO NOT ADD ANY OPTIONS TO IT
\usepackage{algorithm}
\usepackage{newfloat}
\usepackage{listings}
\DeclareCaptionStyle{ruled}{labelfont=normalfont,labelsep=colon,strut=off} % DO NOT CHANGE THIS
\floatstyle{ruled}
\newfloat{listing}{tb}{lst}{}
\floatname{listing}{Listing}
\usepackage{amsmath}
\usepackage{algpseudocode}
\usepackage{booktabs}
\usepackage{subcaption}
\usepackage{makecell}
\usepackage{amsfonts}

\title{MAISI-v2: Accelerated 3D High-Resolution Medical Image Synthesis with Rectified Flow and Region-specific Contrastive Loss}

\author{%
  Can Zhao\textsuperscript{\rm 1},
  Pengfei Guo\textsuperscript{\rm 1},
  Dong Yang\textsuperscript{\rm 1},
  Yucheng Tang\textsuperscript{\rm 1},
  Yufan He\textsuperscript{\rm 1},
  Benjamin Simon\textsuperscript{\rm 2}\textsuperscript{\rm 3},
  Mason Belue\textsuperscript{\rm 4},
  Stephanie Harmon\textsuperscript{\rm 2},
  Baris Turkbey\textsuperscript{\rm 2},
  Daguang Xu\textsuperscript{\rm 1}
}
\affiliations {
    \textsuperscript{\rm 1}NVIDIA,
    \textsuperscript{\rm 2}National Institutes of Health,
    \textsuperscript{\rm 3}University of Oxford,
    \textsuperscript{\rm 4}University of Arkansas for Medical Sciences\\
    canz@nvidia.com
}

\begin{document}
\nocopyright

\maketitle

\begin{abstract}
  Medical image synthesis is an important topic for both clinical and research applications. Recently, diffusion models have become a leading approach in this area. Despite their strengths, many existing methods struggle with (1) limited generalizability that only work for specific body regions or voxel spacings, (2) slow inference, which is a common issue for diffusion models, and (3) weak alignment with input conditions, which is a critical issue for medical imaging. MAISI, a previously proposed framework, addresses generalizability issues but still suffers from slow inference and limited condition consistency. In this work, we present \textbf{MAISI-v2}, the \textbf{first} accelerated 3D medical image synthesis framework that integrates rectified flow to enable fast and high quality generation. To further enhance condition fidelity, we introduce a \textbf{novel} region-specific contrastive loss to enhance the sensitivity to region of interest. Our experiments show that MAISI-v2 can achieve SOTA image quality with $\mathbf{33 \times}$ \textbf{acceleration} for latent diffusion model. We also conducted a downstream segmentation experiment to show that the synthetic images can be used for data augmentation. We release our code, training details, model weights, and a GUI demo to facilitate reproducibility and promote further development within the community.
\end{abstract}

\begin{links}
    \link{Code}{https://github.com/NVIDIA-Medtech/NV-Generate-CTMR/tree/main}
    \link{GUI demo}{https://build.nvidia.com/nvidia/maisi}
\end{links}

\section{Introduction}

Medical image synthesis is a foundational task in medical AI, supporting applications such as data augmentation, modality translation, anomaly simulation, and privacy-preserving data sharing. By generating anatomically plausible and condition-aware 3D images, synthesis models help train robust models in low-data regimes, support cross-modality translation, and enable reproducible benchmarking on synthetic datasets. As medical imaging advances toward more personalized, heterogeneous, and multimodal data, the need for generalizable and controllable synthesis frameworks becomes increasingly critical.

Most existing medical image synthesis methods generally fall into three categories: (1) \textit{direct regression}, which directly predicts voxel intensities from condition inputs using supervised loss functions~\cite{zhao2017whole,dewey2019deepharmony}; (2) \textit{GAN-based models}, which generate realistic images by adversarial training~\cite{yang2020unsupervised,nie2017medical,chartsias2017adversarial,sun2022hierarchical}; and (3) \textit{diffusion-based models}, which learn to reverse a forward noising process through a series of denoising steps~\cite{dorjsembe2024conditional, pinaya2022brain, guo2025maisi,zhu2023make, friedrich2024wdm,chen2024towards,wang20243d,zhu2024generative,wang20253d}. 

Among these three categories, diffusion models have emerged as the state of the art in image generation, due to their stability and ability to model complex distributions. Their application to 3D medical imaging is particularly promising, yet their clinical deployment remains limited due to three primary challenges: (1) large variation for real-world medical images, (2) slow inference for diffusion models like Denoising Diffusion Probabilistic Models~(DDPM)~\cite{ho2020denoising}, and (3) weak condition fidelity.

First, many diffusion-based medical synthesis models are trained on narrow datasets—specific organs, modalities, or voxel spacings—limiting their ability to generalize to real-world clinical data, which varies widely in resolution, anatomy, and acquisition protocol~\cite{dorjsembe2024conditional,pinaya2022brain,zhu2023make,friedrich2024wdm}. Second, diffusion models typically require hundreds of iterative denoising steps during inference, which becomes computationally prohibitive in 3D. Existing methods either can only generate relatively small volumes like $256 \times 256 \times 256$~\cite{dorjsembe2024conditional,pinaya2022brain,zhu2023make,friedrich2024wdm,chen2024towards,xu2024medsyn} or take a long time ~(over 10 minutes) to generate high-resolution volumes such as $512 \times 512 \times 512$~\cite{guo2025maisi,wang20243d}. If the cost to generate a synthetic image is too high, the practical value of medical image synthesis will be limited. Third, while recent work in 2D natural image synthesis like ControlNet++~\cite{li2024controlnetpp} has improved condition fidelity over ControlNet~\cite{zhang2023adding}, such strategies are underexplored in 3D medical synthesis, often leading to unsatisfying alignment between the conditioning input and the output. This is a particularly critical issue in medical image applications.

To address generalization, the MAISI framework~\cite{guo2025maisi} and 3D MedDiffusion~\cite{chen2024towards} were introduced as unified 3D medical image synthesis systems that leverage latent diffusion models (LDMs) and ControlNet to handle diverse anatomical structures and voxel spacings. However, they both still inherit two major limitations of standard LDM and ControlNet: slow inference and lack of robust condition control.

To tackle efficiency, rectified flow models~\cite{liu2022flow} have recently emerged as a powerful alternative to traditional diffusion models. By replacing the stochastic denoising process with a deterministic ordinary differential equation (ODE), rectified flow enables faster sampling while preserving generation quality. This approach has shown promising results in natural image generation and video generation, including Open Sora~\cite{peng2025open} and Stable Diffusion 3.0~\cite{esser2024scaling}, but has not yet been extended to 3D medical imaging.

To improve condition fidelity, cycle-consistency loss like ControlNet++~\cite{li2024controlnetpp} has been proposed in 2D generative models. But it requires to train an additional reverse network using ground-truth conditions, which makes the procedure complicated. Also the error in the reverse network propagates.

In this work, we present \textbf{MAISI-v2}, a next-generation framework for accelerated and condition-consistent 3D medical image synthesis. Our contributions are as follows:

\begin{itemize}
    \item \textbf{First Rectified Flow Model for 3D Medical Image Synthesis:} We adapt rectified flow to the 3D medical domain and integrate it into a latent diffusion framework, achieving substantial $\mathbf{33 \times}$ \textbf{acceleration} in inference while preserving image quality.

    \item \textbf{Novel Region-Specific Contrastive Loss:} We propose a region-specific contrastive loss to enhance the model's sensitivity to small, localized conditions, such as small tumors, while maintaining background consistency.

    \item \textbf{Open-Source Framework and Pretrained Models:} We release code, training details, model weights, and a GUI demo to support reproducibility and provide a strong baseline for future research in controllable and efficient 3D image synthesis.
\end{itemize}

\section{Related Work}

\subsection{MAISI: Medical AI for Synthetic Imaging}
\label{r:maisi}

MAISI~\cite{guo2025maisi} is a modular 3D medical image synthesis framework designed to handle diverse anatomical structures and acquisition protocols. It consists of three key components:

\begin{enumerate}
\item \textbf{Variational Autoencoder (VAE):} MAISI compresses high-resolution 3D volumes into a compact latent space using a VAE trained with a reconstruction loss. The 1-channel images are spatially compressed by $4 \times 4 \times 4$ and generate 4 channel latent features. The overall compression rate is 16.
\item \textbf{DDPM-based Latent Diffusion Model (LDM):} MAISI uses a DDPM-based LDM to generate synthetic latent volumes. Given a latent $z$, the model learns to denoise noisy versions $z_t$ over time steps $t$. The training objective minimizes the standard denoising loss:
\begin{equation}
\mathcal{L}_{\text{ddpm}} = \mathbb{E}_{z, \epsilon, t} \left[ \left\| \epsilon - \epsilon_\theta(z_t, c, t) \right\|^2 \right]
\label{eq:ldm_loss}
\end{equation}
where $c$ includes body region index and voxel spacing conditioning inputs.

\item \textbf{ControlNet:} To provide fine-grained structural guidance, MAISI incorporates ControlNet modules that encode the condition mask and inject it into the LDM through cross-attention or feature modulation. This allows the generation process to be explicitly steered toward the desired anatomy.
\end{enumerate}

Despite its flexibility, MAISI still suffers from slow inference due to DDPM sampling and lacks mechanisms to enforce precise condition fidelity during generation.

\subsection{Rectified Flow}

Rectified Flow is a generative model that learns transport maps between two probability distributions by modeling straight, deterministic paths, referred to as "flows". In theory, there are infinitely many possible paths to map a source distribution $\pi_0$ to a target distribution $\pi_1$. Traditional diffusion models rely on stochastic processes, which typically result in curved or noisy trajectories, requiring many iterative steps during inference to follow such a path from noise to data. 

In contrast, Rectified Flow formulates the transport as a deterministic ordinary differential equation (ODE), encouraging the flow trajectory to be as straight as possible. This substantially reduces the number of required inference steps and accelerates the generation process. It has been adopted in several state-of-the-art (SOTA) frameworks, including Stable Diffusion 3~\cite{esser2024scaling} and Open-Sora~\cite{peng2025open}.

Given samples $x_0 \sim \pi_0$ and $x_1 \sim \pi_1$, Rectified Flow aims to learn a time-dependent velocity field $v_t(x)$ such that the solution $\phi_t(x)$ to the ODE:

\begin{equation}
\frac{d\phi_t(x)}{dt} = v_t(\phi_t(x)), \quad \phi_0(x) = x_0, \quad \phi_1(x) = x_1
\label{eq:rectified_flow_ode}
\end{equation}

The model is trained to minimize the deviation of $v_t(x)$ from the ideal straight-line displacement between $x_0$ and $x_1$ over time. Specifically, the optimization objective is:

\begin{equation}
\mathcal{L}_{\text{flow}} = \int_0^1 \mathbb{E}_{x_0, x_1, t} \left[ \left\| v_t(x_t, c) - (x_1 - x_0) \right\|^2 \right] dt
\label{eq:rectified_flow_loss}
\end{equation}

where $x_t = (1 - t)x_0 + t x_1$ denotes the linear interpolation between the two endpoints. By encouraging $v_t(x)$ to match the constant velocity of a straight path, the model learns to directly connect $\pi_0$ and $\pi_1$ in a more efficient manner, enabling high-quality image synthesis with significantly fewer steps than traditional diffusion methods.

\section{Methodology}
\label{method}

\subsection{Overview of MAISI-v2}

MAISI-v2 builds on the MAISI architecture, by replacing the DDPM-based LDM with a rectified flow model and introducing two loss functions that improve condition fidelity. The training pipeline is similar to MAISI, which is already described in \textbf{Related Work}, consists of:

\begin{itemize}
\item \textbf{VAE:} We reuse the pretrained MAISI VAE to compress 3D volumes without fine-tuning.
\item \textbf{Rectified Flow-based LDM:} We replace the original DDPM with a rectified flow model to accelerate generation, conditioned on the voxel spacing.
\item \textbf{ControlNet with Region-specific Contrastive Loss:} The control branch encodes condition masks and guides synthesis through modulation. It is trained with added region-specific contrastive losses, conditioned on the voxel spacing and the segmentation mask.
\end{itemize}

\subsection{Training Data}
MAISI~\cite{guo2025maisi}, comprising over 10k CT scans. As in MAISI, all images were resampled to dimensions divisible by 128, including 128, 256, 384, 512, and 768 in each axis.
There are three differences compared to the original MAISI training data:
\textbf{(1)} We incorporated the HNSCC dataset~\cite{grossberg2020hnscc} to improve coverage of the head-and-neck region, as well as an additional in-house dataset containing 1.4k chest and abdomen CT scans. Together, these additions increased the total dataset size from 10k to 12k scans. \textbf{(2)} We removed CT scans with fewer than 64 slices to ensure data quality, resulting in 11k usable scans. \textbf{(3)} We downsampled high-resolution images to augment the representation of low-resolution cases. The final training set consists of over 107k images generated from the 11k scans.

For ControlNet training, same with MAISI~\cite{guo2025maisi}, we use CT images and the corresponding pseudo segmentation maps with 127 types of human anatomical structures and various lesions, which are obtained from the open-source model VISTA3D~\cite{he2024vista3d}. For some datasets, manual labels are available for a few organs or tumors. In this case, we replace the corresponding pseudo labels with real labels. This setup allows the ControlNet module to learn spatially accurate synthesis conditioned on semantic labels.

\subsection{Rectified Flow-based LDM Training}
\label{m:ldm}
Naively training images with varying shapes all together will cause the batch size to be restricted to 1. And small batch size will make the LDM training process prone to NaN issues for mixed-precision optimization. To improve both stability and convergence speed for training, we developed a three-stage training strategy with mixed-precision optimization. The only conditioning signal is the voxel spacing, which is automatically extracted from the NIfTI header and requires no manual annotation. Training is conducted on 64 NVIDIA A100 GPUs (80GB) over three weeks using the AdamW optimizer.

The three training stages are outlined below.

\textbf{Pre-training Stage:} In the first stage, we use low-resolution images of size $128 \times 128 \times 128$, consisting of both naturally low-resolution volumes and high-resolution images that have been downsampled. Since all images share the same shape, we are able to use a large batch size of 96 on 80GB A100 GPUs. This stage is critical as it not only accelerates training significantly but also helps prevent NaN issues that may arise in subsequent stages. We use a learning rate of 1e-3 to train this stage for one day.

\textbf{Main Stage:} Once the low-resolution model converges, we move to the second stage and continue training on the full-resolution dataset, where images vary in spatial dimensions. To avoid using batch size as 1, we use bucketed data parallelism. We group images by shape and distribute them to separate GPUs, ensuring that each GPU processes only images of the same shape, and then train the model with distributed data parallel. This allows us to use larger batch sizes where possible. For example, $128 \times 128 \times 128$ images are trained with a batch size of 96, $256 \times 256 \times 128$ with a batch size of 24, and $512 \times 512 \times 768$ with a batch size of 1. We use a learning rate of 1e-3 and polynomial decay for this stage for 16000 epochs. It took around 10 days. While this strategy greatly accelerates training, it introduces data imbalance issue.

\textbf{Fine-Tuning Stage:} To address the data imbalance issue from the second stage, we add a third fine-tuning stage, where the model is trained using all images mixed together with a fixed batch size of 1 and sampling weights to balance the contribution from different datasets. This ensures the model is exposed to the full data distribution and helps reduce any size-related bias in the final representation. We use a learning rate of 1e-4 for this stage for around 2000 epochs. It took around 10 days.

\subsection{ControlNet Training}
The ControlNet training was done with 8 A100 GPUs, using AdamW optimizer and a learning rate of 5e-5 and polynomial decay. We trained the network with 60 epochs and it takes around 2 days.

Let $x_0 \sim \pi_0$ be a latent Gaussian noise sample, and $x_1 \sim \pi_1$ be the latent encoding of a real image $I$. The conditioning input $c$ (e.g., segmentation mask) is processed through a separate encoder $E_\theta$ and fused into the frozen diffusion model and form the generator $G_\theta$ which predicts velocity.
\begin{equation}
v_t(x, c) = G_\theta(x_t, c)
\label{eq:rectified_flow}
\end{equation}

We train a time-dependent network $G_\theta(x_t, c)$, conditioned on both $x_t$ and $c$, to minimize the deviation from the linear displacement, as shown in Equation~\ref{eq:rectified_flow_loss}.

In our setup, the conditioning input $c$ is a segmentation mask. To emphasize the response in regions of interest (ROIs), specifically tumors, we applied a weighted average loss with a tumor weight of 100. However, this alone was insufficient to ensure clear tumor appearance in the generated outputs. To address this limitation, we introduce a Region-Specific Contrastive Loss.

\subsubsection{Region-specific Contrastive Loss}

\begin{figure*}[t]
\centering
    \includegraphics[width=0.7\linewidth]{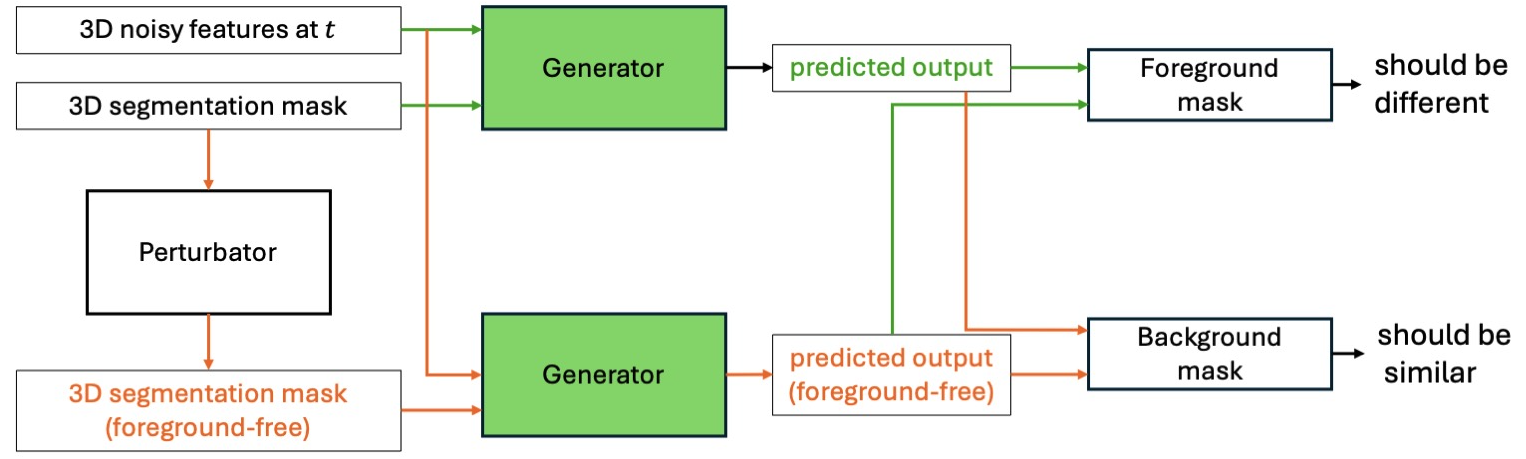}
\caption{Region-specific Contrastive Loss}
\label{fig:loss}
\end{figure*}
To explicitly enforce condition fidelity, we propose a differential conditioning strategy that disentangles the model's response to the region of interest (ROI) versus the background. The key idea is to generate two versions of the output from the same noise input: one conditioned on the original segmentation mask, and another on a perturbed mask where the ROI labels are replaced with the corresponding background labels (e.g., replacing a pancreatic tumor label with a pancreas label). The synthesized images should differ within the ROI (reflecting the conditioning change), while remaining consistent in the background (which is unaltered).

This results in a composite loss with two complementary objectives:  
1) \textbf{ROI Sensitivity Loss}, which encourages discriminative synthesis in regions where the condition differs;  
2) \textbf{Background Consistency Loss}, which enforces invariance in regions where the condition remains unchanged.  
We term this a \textbf{Region-specific Contrastive Loss}—not in the classical embedding sense, but as a mechanism that explicitly contrasts outputs under different conditioning inputs within the ROI.

The high-level idea is presented in Figure~\ref{fig:loss}. The detailed procedure is described as follows. 
\begin{algorithm}[h!]
\caption{Region-Specific Contrastive Loss Computation}
\small
\begin{algorithmic}[1]
\Require Generator $G_\theta$, condition mask $c_{\text{orig}}$, cap $\delta$
\State Sample noise $x_0 \sim \pi_0$
\State Construct perturbed mask $c_{\text{perturb}} \gets c_{\text{orig}}$: replace ROI labels with corresponding background labels, e.g., replacing pancreatic tumor label with pancreas label.
\State Compute ROI binary mask $m \gets \mathbb{1}(c_{\text{orig}} \neq c_{\text{perturb}})$
\State Generate predictions with $c_{\text{orig}}$ and $c_{\text{perturb}}$: $G_\theta(x_t, c_{\text{orig}})$ and $G_\theta(x_t, c_{\text{perturb}})$
\State Compute ROI sensitivity loss $\mathcal{L}_{\text{roi}}$ using Eq.~\eqref{eq:roi_loss}
\State Compute background consistency loss $\mathcal{L}_{\text{bg}}$ using Eq.~\eqref{eq:bg_loss}
\State \Return $\mathcal{L}_{\text{roi}} + \mathcal{L}_{\text{bg}}$
\end{algorithmic}
\end{algorithm}
\begin{align}
\mathcal{D}_{\text{roi}} &= \left\| 
\left( G_\theta(x_t, c_{\text{orig}}) - G_\theta(x_t, c_{\text{perturb}}) \right) 
\odot m \right\|_{1, m} \\
\mathcal{L}_{\text{roi}} &=-\text{min}(\mathcal{D}_{\text{roi}},\delta) \label{eq:roi_loss}\\
m^- &= 1 - \text{dilate}(m)\\
\mathcal{L}_{\text{bg}} &= \left\| (G_\theta(x_t, c_{\text{orig}}) - G_\theta(x_t, c_{\text{perturb}})) \odot m^-) \right\|_{1, m^-} \label{eq:bg_loss}
\end{align}

As shown in Eq.~\eqref{eq:roi_loss}, we apply an upper bound $\delta$ to prevent gradient explosion during training. This is important. Removing this bound leads to unstable optimization and results in NaN values during training. Intuitively, the goal is not to maximize the difference within the ROI indefinitely, but merely to ensure the model can distinguish them. Since the latents were generated with a trained VAE encoder, we assume the values of feature maps approximately follows Normal Distribution. Therefore, we set $\delta = 2$, corresponding to a $\pm$ standard deviation range.

\subsubsection{Memory-Aware Region-specific Contrastive Loss}
When GPU memory is a bottleneck, we need to reduce the memory usage of Region-specific Contrastive Loss for large image inputs. To accommodate varying input sizes and GPU memory constraints, we adaptively select the feature source for loss computation.
\begin{itemize}
    \item For \textbf{small to medium-sized inputs}, we compute the loss on the outputs of the ControlNet and frozen diffusion model $G_\theta$. These final output features provide higher spatial fidelity and stronger alignment with the conditioning mask. This is preferred if GPU memory allows.
    
    \item For \textbf{large inputs}, we apply the loss on the encoded features of the ControlNet encoder $E_\theta$, without passing through the frozen diffusion model. Although these intermediate features are coarser, this strategy enables training on high-resolution volumes within memory limits.
\end{itemize}

In both cases, the loss terms $\mathcal{L}_{\text{roi}}$ and $\mathcal{L}_{\text{bg}}$ remain unchanged. The losses are computed either on the final outputs of the ControlNet and diffusion model, or on intermediate features from the ControlNet encoder, depending on input size and memory availability.

\subsubsection{Final Objective}

The full training objective combines all components:
\begin{equation}
\mathcal{L}_{\text{total}} = \mathcal{L}_{\text{flow}} + \lambda_{\text{contrast}} (\mathcal{L}_{\text{roi}} + \mathcal{L}_{\text{bg}})
\label{eq:loss}
\end{equation}
where $\lambda_{\text{contrast}}$ are hyperparameters controlling the weight of the condition fidelity terms.

\begin{figure*}[t]
\centering

% Row 0: Method Names
\begin{subfigure}{0.195\textwidth}
\centering
\parbox{\linewidth}{\centering \textbf{HA-GAN~\cite{sun2022hierarchical}}}
\end{subfigure}
\begin{subfigure}{0.195\textwidth}
\centering
\parbox{\linewidth}{\centering \textbf{GenerateCT~\cite{hamamci2024generatect}}}
\end{subfigure}
\begin{subfigure}{0.195\textwidth}
\centering
\parbox{\linewidth}{\centering \textbf{MedSyn~\cite{xu2024medsyn}}}
\end{subfigure}
\begin{subfigure}{0.195\textwidth}
\centering
\parbox{\linewidth}{\centering \textbf{MAISI-DDPM~\cite{guo2025maisi}}}
\end{subfigure}
\begin{subfigure}{0.195\textwidth}
\centering
\parbox{\linewidth}{\centering \textbf{MAISI-v2}}
\end{subfigure}

% Row 1: Axial
\begin{subfigure}{0.195\textwidth}
    \includegraphics[width=\linewidth]{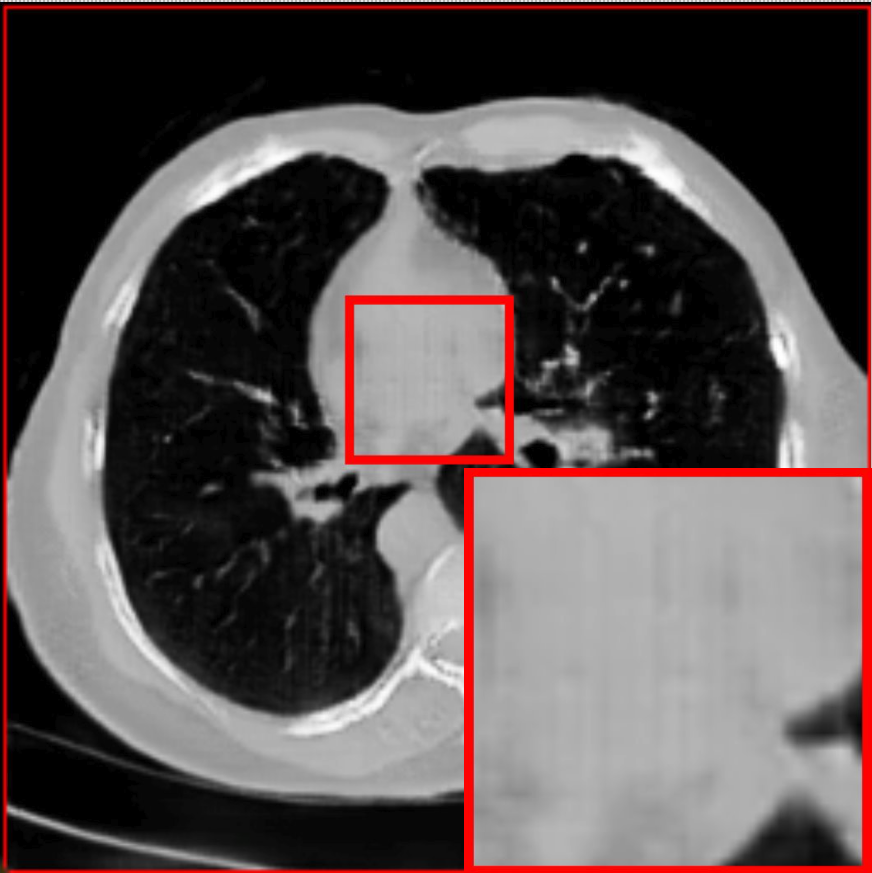}
\end{subfigure}
\begin{subfigure}{0.195\textwidth}
    \includegraphics[width=\linewidth]{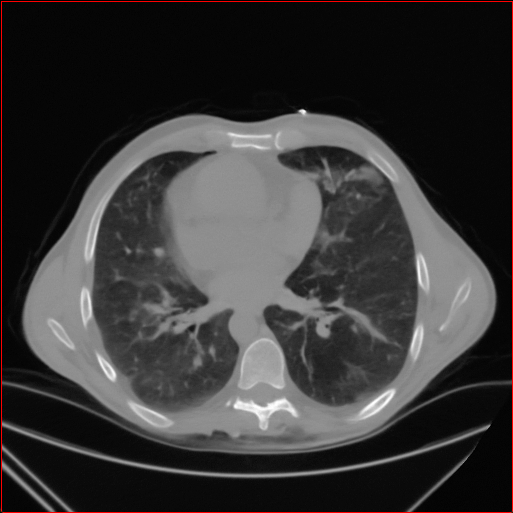}
\end{subfigure}
\begin{subfigure}{0.195\textwidth}
    \includegraphics[width=\linewidth]{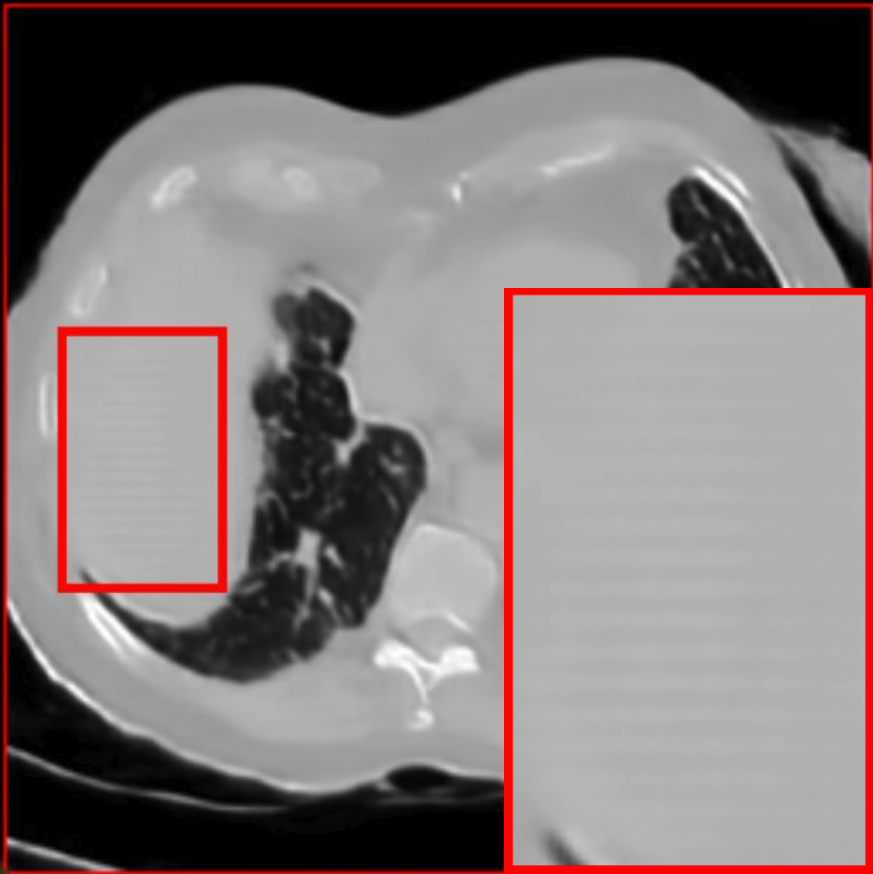}
\end{subfigure}
\begin{subfigure}{0.195\textwidth}
    \includegraphics[width=\linewidth]{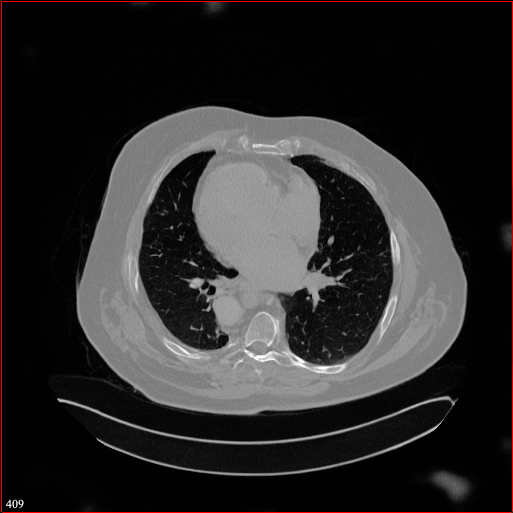}
\end{subfigure}
\begin{subfigure}{0.195\textwidth}
    \includegraphics[width=\linewidth]{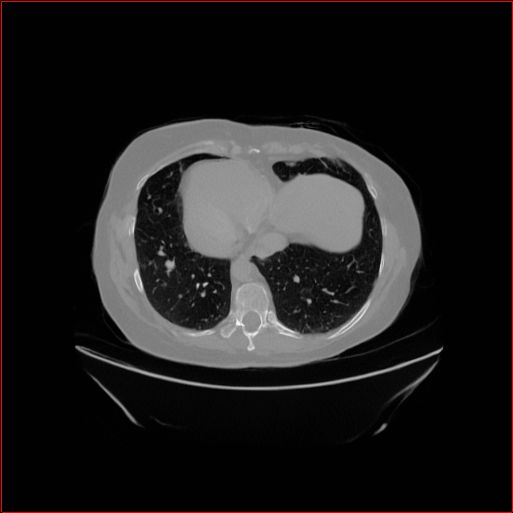}
\end{subfigure}

% Row 2: Sagittal
\begin{subfigure}[c]{0.195\textwidth}
    \includegraphics[width=\linewidth]{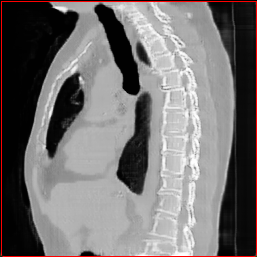}
\end{subfigure}
\begin{subfigure}[c]{0.195\textwidth}
    \includegraphics[width=\linewidth]{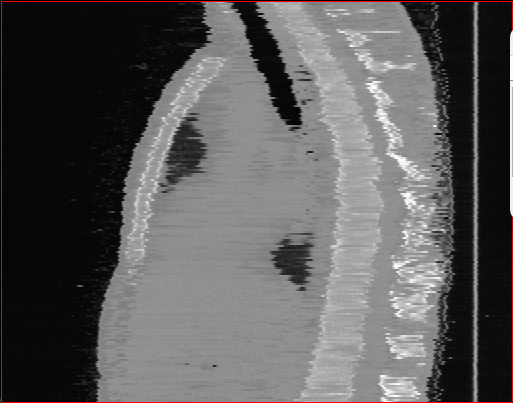}
\end{subfigure}
\begin{subfigure}[c]{0.195\textwidth}
    \includegraphics[width=\linewidth]{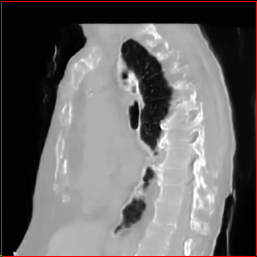}
\end{subfigure}
\begin{subfigure}[c]{0.195\textwidth}
    \includegraphics[width=\linewidth]{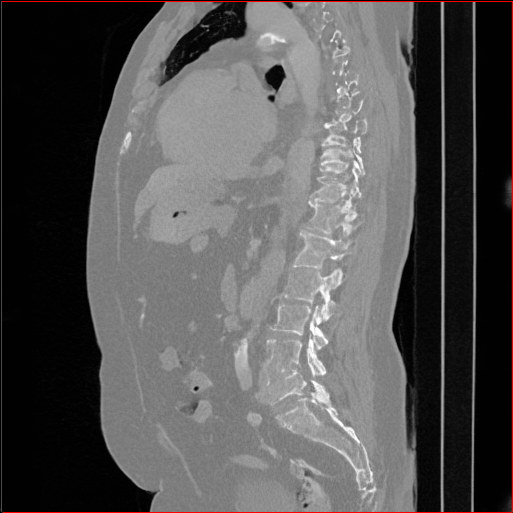}
\end{subfigure}
\begin{subfigure}[c]{0.195\textwidth}
    \includegraphics[width=\linewidth]{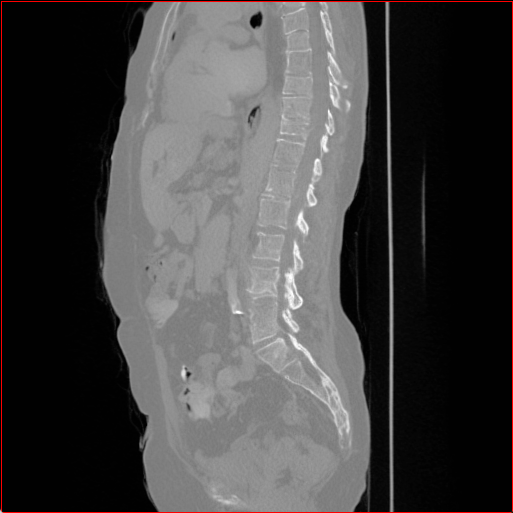}
\end{subfigure}

% Row 3: Coronal
\begin{subfigure}[c]{0.195\textwidth}
    \includegraphics[width=\linewidth]{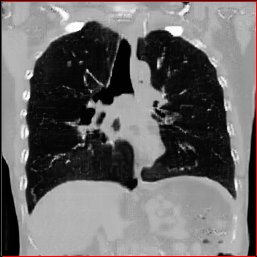}
\end{subfigure}
\begin{subfigure}[c]{0.195\textwidth}
    \includegraphics[width=\linewidth]{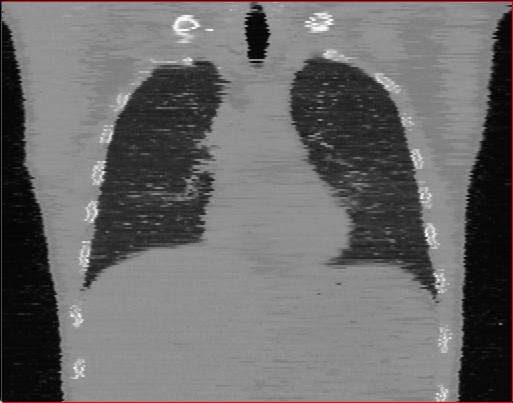}
\end{subfigure}
\begin{subfigure}[c]{0.195\textwidth}
    \includegraphics[width=\linewidth]{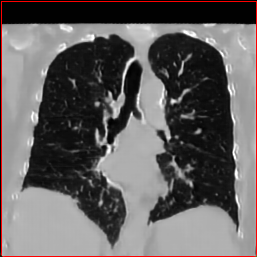}
\end{subfigure}
\begin{subfigure}[c]{0.195\textwidth}
    \includegraphics[width=\linewidth]{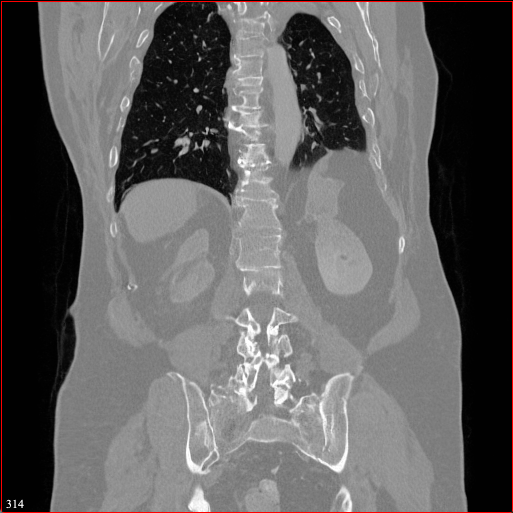}
\end{subfigure}
\begin{subfigure}[c]{0.195\textwidth}
    \includegraphics[width=\linewidth]{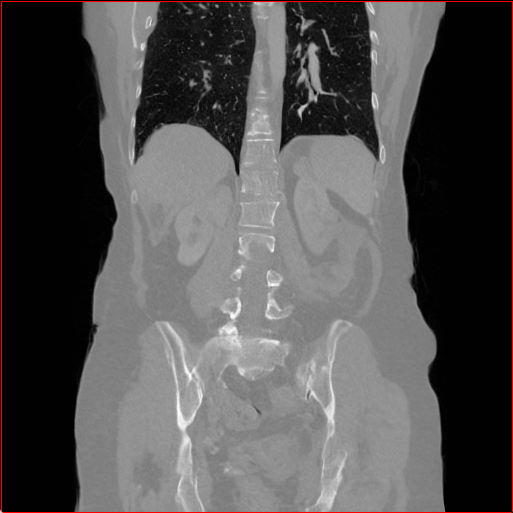}
\end{subfigure}

\caption{Qualitative comparison across axial (top row), sagittal (middle row), and coronal (bottom row) views. Columns correspond to different methods. MAISI-DDPM and MAISI-v2 in this figure are unconditional synthesis which do not use ControlNet or segmentation maps.}
\label{fig:qualitative}
\end{figure*}

\subsubsection{Quality Check with CT HU Intensities}
We performed a quality check on the Hounsfield Unit (HU) intensities of generated CT images. Intuitively, the median HU values of major organs in generated images should fall within realistic physiological ranges. 

To determine these ranges, we collected the median HU values for the liver, spleen, pancreas, kidneys, lungs, bones, and brain from all the training images. With these values, the lower threshold for each organ was set as the more extreme of the 5th percentile or the 6-sigma lower bound, and the upper threshold was defined similarly using the 95th percentile and the 6-sigma upper bound.

This quality check is related to choosing $\lambda_{\text{contrast}}$. Large $\lambda_{\text{contrast}}$ leads to results with tumors well aligned with tumor masks. Yet if too large, the rest of the body structures will be wrong and the result will not be able to pass the quality check. Our criterion to choose $\lambda_{\text{contrast}}$ is a large $\lambda_{\text{contrast}}$ that can make nearly 100\% of the generated images pass the quality check.

\begin{table*}[t]
\small
\centering
\caption{FID scores and single 80G GPU inference time cost on the OOD dataset AutoPET2023 across orthogonal planes and overall average (lower is better). All synthetic images are $512 \times 512 \times 512$ volumes with voxel spacing of $1 \times 1 \times 1$~$\textrm{mm}^3$. For MAISI and MAISI-v2, the time cost includes 30 steps LDM (6s) or 1000 steps LDM (198s), and VAE decoding (15s) on H100.}
\label{tab:fid}
\begin{tabular}{lcccccccc}
\toprule
\textbf{Model} & \textbf{Steps} & \textbf{Time (s)} & $\text{FID}_{xy}$ & $\text{FID}_{yz}$ & $\text{FID}_{xz}$ & $\text{FID}_{avg}$ & \textbf{Device} \\
\midrule
HA-GAN & 1 & ~1  & 13.813 & 12.567 & 14.405 & 13.595 & H100 \\
MedSyn (2-stage DDIM) & 50+20 & 100  & 18.662 & 22.171 & 33.293 & 24.709 & H100 \\
GenerateCT (2D EDM) & 25$\times$201 & 89 & 7.909 & 9.256 & 15.106 & 10.757 & H100 \\
MAISI (DDPM) & 1000 & 198+15  & \textbf{2.199} & 2.480 & 2.642 & 2.441 & H100 \\
MAISI (DDIM) & 30 & 6+15  & 4.855 & 4.703 &4.770  & 4.776 & H100 \\
MAISI-v2 (Rectified Flow) & 30 & 6+15 &  2.217 & \textbf{2.211} & \textbf{2.538} & \textbf{2.322} & H100 \\
\bottomrule
\end{tabular}
\end{table*}

\begin{table*}[h]
\small
\centering
\caption{Inference time comparison across different methods and image/video sizes on single 80G GPU. MAISI-v2 (Unconditioned) refers to the time cost of LDM and VAE decoding. MAISI-v2 (Conditioned) refers to the time cost of ControlNet, LDM, and VAE decoding. All diffusion models in this table have 50 inference steps. The devices are different since we used the officially reported time cost from the original papers.}
\label{tab:time_comparison}
\begin{tabular}{lcccc}
\toprule
\textbf{Method (50 steps inference)} & \textbf{Output Size} & \textbf{\# Voxels} & \textbf{Time (s)}& \textbf{Device} \\
\midrule
SVD~\cite{blattmann2023stable} (Conditioned) & $3 \times 576 \times 1024 \times 25$ (video) & 4.4E7 & 100 & A100 \\
Open Sora 2.0~\cite{peng2025open} (Conditioned)& $3 \times 768 \times 768 \times 128$ (video) & 2.3E8& 162 & H200 \\
MAISI-v2 (Unconditioned) & $512 \times 512 \times 512$ (image) & 1.3E8 & 26 & H100\\
MAISI-v2 (Conditioned) & $512 \times 512 \times 512$ (image) & 1.3E8 & 34 & H100\\
\bottomrule
\end{tabular}
\end{table*}

\section{Experiments}
\label{e}
We implement all networks using PyTorch~\cite{ansel2024pytorch} and MONAI~\cite{cardoso2022monai}. All the evaluation experiments were done with 80G H100 GPUs. 

\subsection{Training Data and Implementation Details}
The training dataset consists of many public datasets including HNSCC dataset~\cite{grossberg2020hnscc}, AbdomenCT-1K~\cite{ma2021abdomenct}, AeroPath ~\cite{stoverud2023aeropath}, AMOS22~\cite{ji2022amos}, COVID-19~\cite{chowdhury2020can,rahman2021exploring}, CRLM-CT~\cite{simpson2024preoperative}, CT-ORG~\cite{rister2019ct}, CTPelvic1K-CLINIC~\cite{liu2021deep}, LIDC~\cite{armato2011lung}, MSD Task03, Task06, Task07, Task08, Task09, Task10~\cite{antonelli2022medical}, Multi-organ-Abdominal-CT~(BTCV, Pancreas-CT)~\cite{gibson2018automatic,landman2015segmentation,roth2016data}, NLST~\cite{national2011reduced}, StonyBrook-CT~\cite{saltz2021stony}, TCIA Colon~\cite{johnson2008accuracy}, TotalSegmentatorV2~\cite{wasserthal2023dataset}. We also added two in-house datasets, including 237 bone lesion CT and 1386 chest\&abdomen CT to further enrich the training dataset, especially for CT images with bone lesions and corresponding masks.

For implementation details in ControlNet, we found that large $\lambda_{\text{contrast}}$ leads to results with tumors well aligned with masks. Yet if too large, the rest of the body structures will be wrong. 

In order to tackle this issue, for the 60 training epochs, we empirically choose $\lambda_{\text{contrast}}=0.01$ for the first 30 epochs to make sure the tumors appear in the generated results, then decrease to $\lambda_{\text{contrast}}=0.001$ to make the rest of the body structure correct.
We experimented with the opposite way, i.e., first use small $\lambda_{\text{contrast}}$ then increase, the tumors did not show up as expected. In the final models, 100\% of the results passed the quality check on CT HU intensities.

\subsection{Evaluation of Accelerated LDM}
\label{e:ldm}
We first evaluate the performance of Rectified Flow LDM, without involving the ControlNet.

For comparison methods, to the best of our knowledge, the only methods that can also generate high-resolution 3D CT images with size equal or larger than $512 \times 512 \times 512$ are MAISI~\cite{guo2025maisi} and 3D MedDiffusion~\cite{wang20243d}, while only MAISI is open-sourced. For MAISI and MAISI-v2, they supports various of size and voxel spacing. We generate synthetic images with size of $512 \times 512 \times 512$ and voxel spacing of $1 \times 1 \times 1$~$\textrm{mm}^3$ for evaluation purpose to show its ability to cover a large body region. Other methods generate images with smaller volume size. HA-GAN~\cite{sun2022hierarchical} generates images with a fixed size of $256 \times 256 \times 256$ and unknown voxel spacing as their training data were zoomed to $256 \times 256 \times 256$ without considering the voxel spacing. MedSyn~\cite{xu2024medsyn} generates images with a fixed size of $256 \times 256 \times 256$ and fixed voxel spacing of $1 \times 1 \times 1$~$\textrm{mm}^3$. GenerateCT~\cite{hamamci2024generatect} generates images with a fixed size of $512 \times 512 \times 201$ and fixed voxel spacing of $0.75 \times 0.75 \times1.5$~$\textrm{mm}^3$. GenerateCT first generates low-resolution 3D volumes, then applies 2D slice-by-slice upscaling, which causes its low-quality in sagittal and coronal planes. 

\paragraph{Qualitative Evaluation:} Figure~\ref{fig:qualitative} presents representative slices from the axial, sagittal, and coronal planes. GenerateCT~\cite{hamamci2024generatect} is a 2D model, so it lacks inter-slice consistency, leading to poor image quality in the sagittal and coronal views. MedSyn~\cite{xu2024medsyn} produces noticeably blurry results with mosaic-like artifacts, such as region inside the red box. HA-GAN~\cite{sun2022hierarchical} generates visually sharp images but with mosaic-like artifacts, such as region inside the red box. Also, its voxel spacing is not available, which limits its applicability in real-world medical imaging tasks. Moreover, all three methods are restricted to synthesizing small anatomical regions. In contrast, both MAISI and MAISI-v2 are capable of generating high-quality 3D volumes that span larger body regions while preserving fine anatomical details and realistic structure.

\paragraph{Quantitative Evaluation:} we use the AutoPET2023 dataset~\cite{gatidis2022whole} as an out-of-distribution (OOD) benchmark to evaluate their generalizability.

For OOD dataset AutoPET2023, it contains a wide range of image resolution. In order to evaluate high-resolution image synthesis, we select original AutoPET2023 images with voxel spacing smaller than $1 \times 1 \times 1$~$\textrm{mm}^3$ and FOV larger than $512 \times 512 \times 512$~$\textrm{mm}^3$, then downsample them to $1 \times 1 \times 1$~$\textrm{mm}^3$ and central crop $512 \times 512 \times 512$ volumes.

For MAISI and MAISI-v2, We generate synthetic images with size of $512 \times 512 \times 512$ and voxel spacing of $1 \times 1 \times 1$~$\textrm{mm}^3$. For other comparison methods, the size of the generated images is fixed. We resampled them to $1 \times 1 \times 1$~$\textrm{mm}^3$ and zero-padded to $512 \times 512 \times 512$. MAISI was trained with 1000 step DDPM. But there are schedulers that can accelerate inference like Denoising Diffusion Implicit Models~(DDIM)~\cite{song2020denoising} and Elucidated Diffusion Sampler~(EDM)~\cite{karras2022elucidating}. We tested DDIM scheduler for MAISI. 

The metric we use is 2D Fréchet Inception Distance (FID) between synthesized images and the unseen AutoPET2023 dataset for $xy$, $yz$, and $xz$-planes. The FID score measures how closely the distribution of generated samples matches the real data distribution. Lower values indicate higher fidelity and better generalization. To avoid empty slices, we compute FID on the central 50\% of the slices. The reason we use 2D FID in three planes instead of 3D FID is to better reflect human perception.

Table~\ref{tab:fid} summarizes the results. It shows that MAISI-v2 achieves at least comparable FID with MAISI (DDPM) with much faster speed. For MAISI and MAISI-v2, the time cost in Table~\ref{tab:fid} is the total of the LDM and VAE decoding time cost. The LDM generates $4 \times 128 \times 128 \times 128$ latent features. They are then decoded by MAISI VAE, which takes 15s to generate a $512 \times 512 \times 512$ volume on H100 using a sliding window patch size of $80 \times 80 \times 80$ on latent features. 

While Table~\ref{tab:fid} compares MAISI-v2’s inference speed with other medical image generation methods, Table~\ref{tab:time_comparison} extends the comparison to SOTA video generation models, and provides additional perspective on the LDM inference efficiency across domains. Specifically, we include Stable Video Diffusion (SVD)~\cite{blattmann2023stable} and Open-Sora 2.0~\cite{peng2025open}, both of which use 50 inference steps. To ensure consistency, we also report MAISI-v2’s runtime using 50 inference steps.

\begin{table*}[!htbp]
\centering
\caption{FID scores across inference steps for MAISI-v2 LDM. Each value represents the score on the OOD AutoPET23 dataset (lower is better).}
\label{tab:fid_full}
\begin{tabular}{lcccccccc}
\toprule
\textbf{Metric} & \textbf{5} & \textbf{10} & \textbf{20} & \textbf{30} & \textbf{40} & \textbf{50} & \textbf{70} & \textbf{100} \\
\midrule
FID$_{xy}$  & 22.774 & 4.232 & 2.538 & 2.217 & 2.088 & 1.966 & 1.818 & 1.783 \\
FID$_{yz}$  & 13.907 & 3.765 & 2.446 & 2.211 & 2.119 & 2.001 & 1.970 & 1.960 \\
FID$_{xz}$  & 24.321 & 5.266 & 2.952 & 2.538 & 2.405 & 2.225 & 2.156 & 2.159 \\
FID$_{avg}$ & 20.334 & 4.421 & 2.645 & 2.322 & 2.204 & 2.064 & 1.981 & 1.967 \\
\bottomrule
\end{tabular}
\end{table*}

\paragraph{Ablation Study:} We conduct an ablation study to analyze how the number of inference steps affects the quality of images synthesized by the rectified flow-based MAISI-v2 model. Same as Table~\ref{tab:fid}, FID with the same AutoPET23 reference images is computed along three orthogonal planes ($xy$, $yz$, and $xz$), as well as the averaged score (FID$_\text{avg}$).

As shown in Table~\ref{tab:fid_full}, image quality improves consistently as the number of inference steps increases from 5 to 100, with diminishing returns beyond 30 steps. 

Based on this analysis, we choose 30 inference steps as the default setting for most experiments in this paper. This setting offers a favorable balance between synthesis quality and computational efficiency.

\subsection{Evaluation of ControlNet with Additional Losses}
\label{e:controlnet}
\begin{figure*}[!htbp]
\centering
\begin{subfigure}{0.32\textwidth}\centering \textbf{Head Region}\\ \small{$1.1 \times1.1 \times 1.1$~mm \\ $256 \times256 \times 256$}\end{subfigure}
\begin{subfigure}{0.32\textwidth}\centering \textbf{Chest Region}\\ \small{$1 \times1 \times 1$~mm \\ $384 \times384 \times 384$}\end{subfigure}
\begin{subfigure}{0.32\textwidth}\centering \textbf{Abdomen Region}\\ \small{$1 \times1 \times 0.7$~mm \\ $512 \times512 \times 768$}\end{subfigure}

% Row 1: Axial
\begin{subfigure}{0.3\textwidth}
\includegraphics[width=\linewidth]{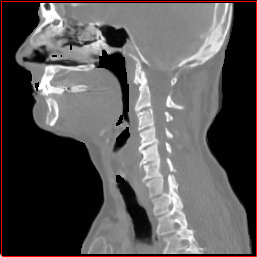}
\end{subfigure}
\begin{subfigure}{0.3\textwidth}
    \includegraphics[width=\linewidth]{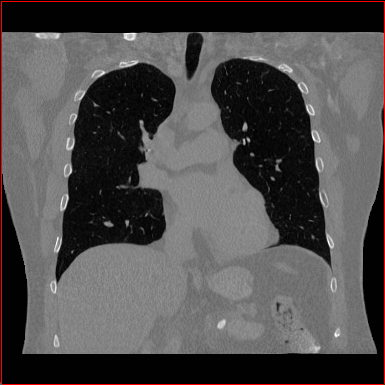}
\end{subfigure}
\begin{subfigure}{0.3\textwidth}
    \includegraphics[width=\linewidth]{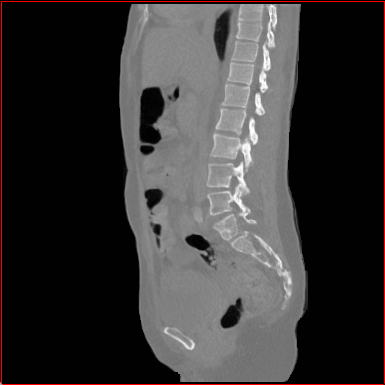}
\end{subfigure}
\caption{MAISI-v2 segmentation-guided results for small to large volume size and three different regions.}
\label{fig:size}
\end{figure*}

\begin{figure*}[!htbp]
\centering
% Row 0: Method Names
\begin{subfigure}{0.195\textwidth}\centering \textbf{Lung Tumor}\\ \small{$0.75 \times0.75 \times 0.6$~mm \\ $512 \times512 \times 512$}\end{subfigure}
\begin{subfigure}{0.195\textwidth}\centering \textbf{Liver Tumor}\\ \small{$0.75 \times0.75 \times 0.5$~mm \\ $512 \times512 \times 768$}\end{subfigure}
\begin{subfigure}{0.195\textwidth}\centering \textbf{Panc. Tumor}\\ \small{$1 \times1 \times 1$~mm \\ $512 \times512 \times 512$}\end{subfigure}
\begin{subfigure}{0.195\textwidth}\centering \textbf{Colon Tumor}\\ \small{$0.75 \times0.75\times 1.5$~mm \\ $512 \times512 \times 256$}\end{subfigure}
\begin{subfigure}{0.195\textwidth}\centering \textbf{Bone-Les}\\ \small{$1 \times1 \times 1.3$~mm \\ $512 \times512 \times 384$}\end{subfigure}

% Row 1: Axial
\begin{subfigure}{0.195\textwidth}
\includegraphics[width=\linewidth]{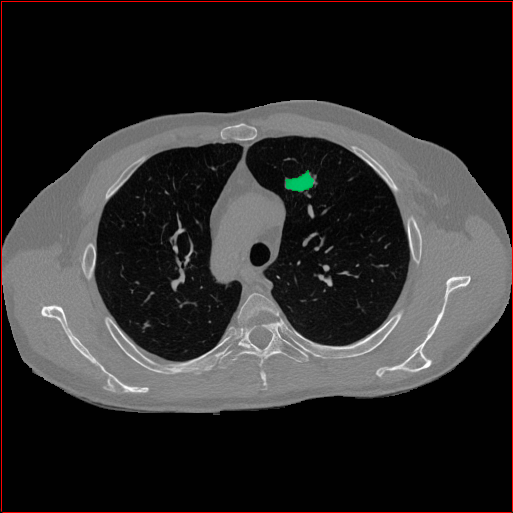}
\end{subfigure}
\begin{subfigure}{0.195\textwidth}
    \includegraphics[width=\linewidth]{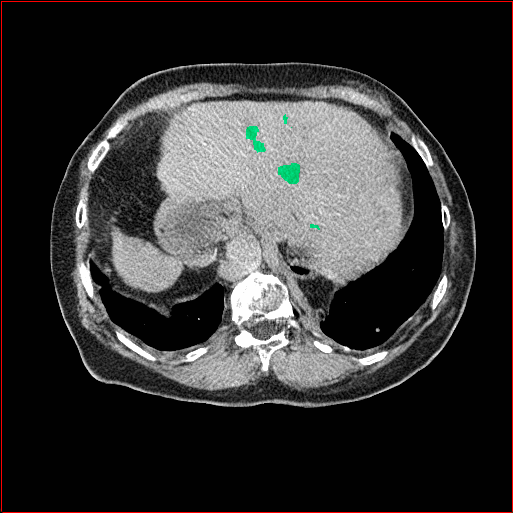}
\end{subfigure}
\begin{subfigure}{0.195\textwidth}
\includegraphics[width=\linewidth]{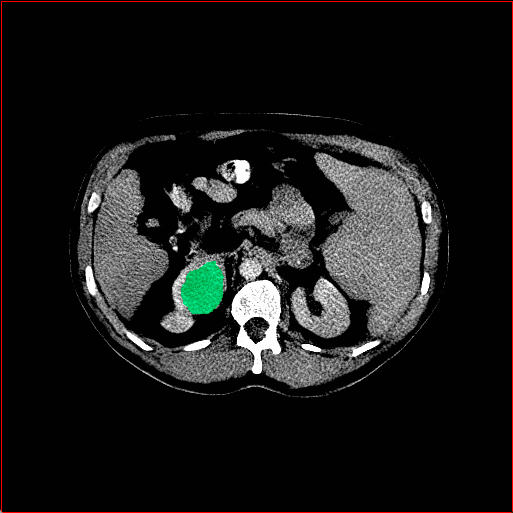}
\end{subfigure}
\begin{subfigure}{0.195\textwidth}
    \includegraphics[width=\linewidth]{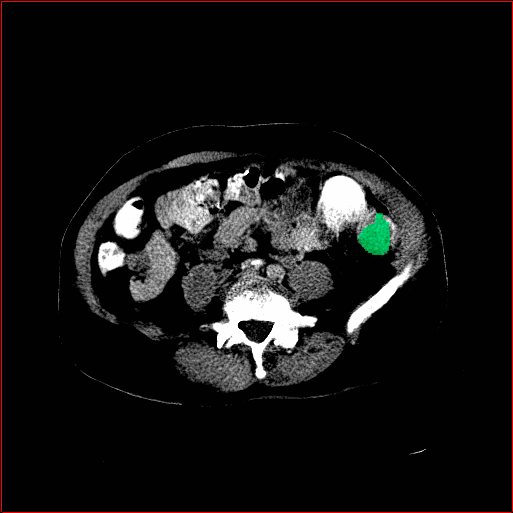}
\end{subfigure}
\begin{subfigure}{0.195\textwidth}
    \includegraphics[width=\linewidth]{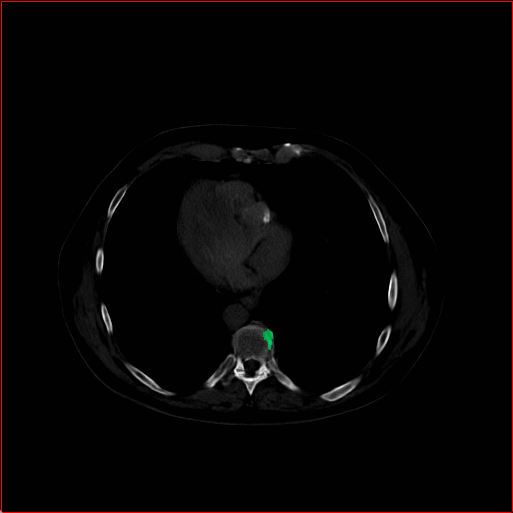}
\end{subfigure}

\begin{subfigure}[c]{0.195\textwidth}
    \includegraphics[width=\linewidth]{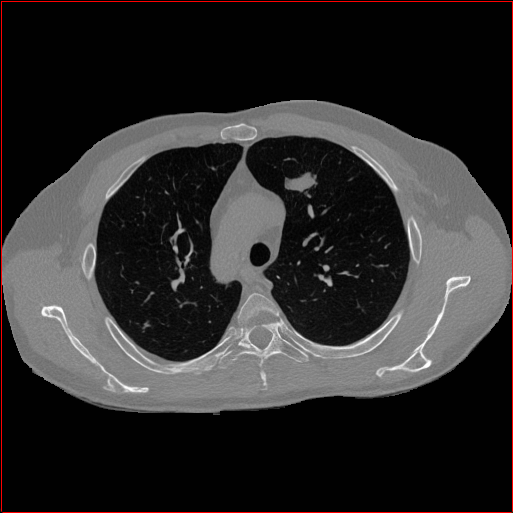}
\end{subfigure}
\begin{subfigure}[c]{0.195\textwidth}
    \includegraphics[width=\linewidth]{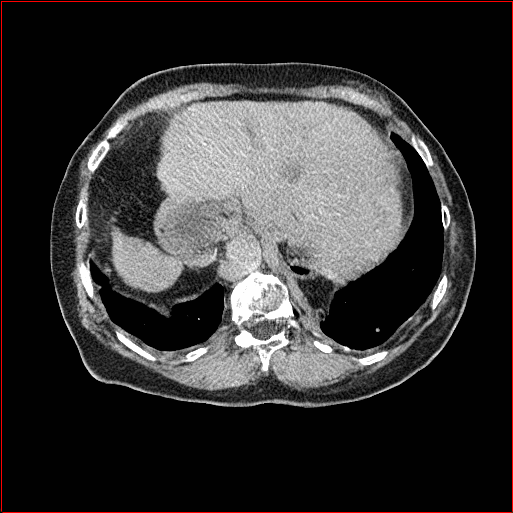}
\end{subfigure}
\begin{subfigure}[c]{0.195\textwidth}
    \includegraphics[width=\linewidth]{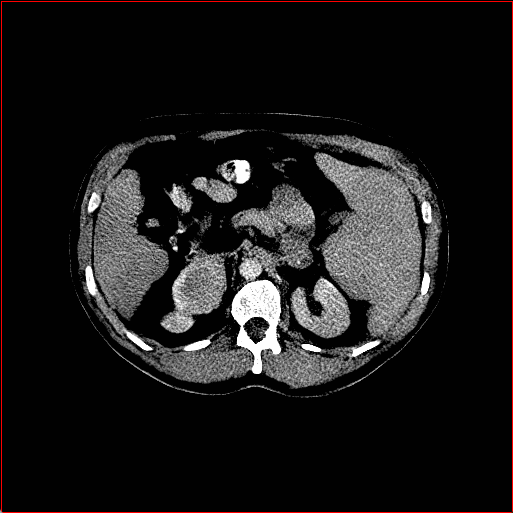}
\end{subfigure}
\begin{subfigure}[c]{0.195\textwidth}
    \includegraphics[width=\linewidth]{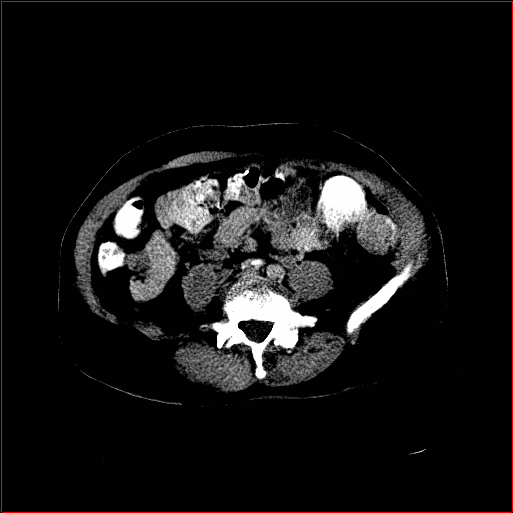}
\end{subfigure}
\begin{subfigure}[c]{0.195\textwidth}
    \includegraphics[width=\linewidth]{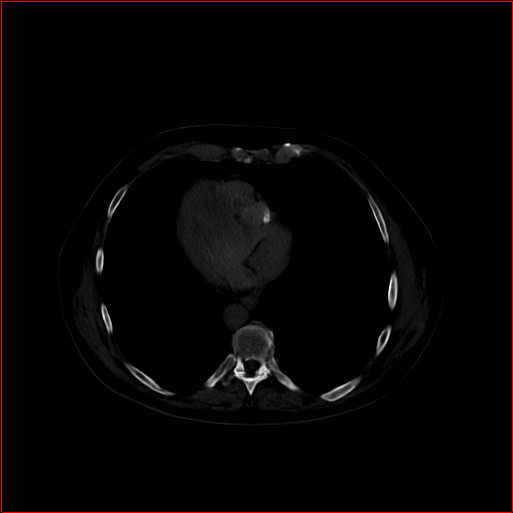}
\end{subfigure}

% Row 2: Sagittal
\begin{subfigure}[c]{0.195\textwidth}
    \includegraphics[width=\linewidth]{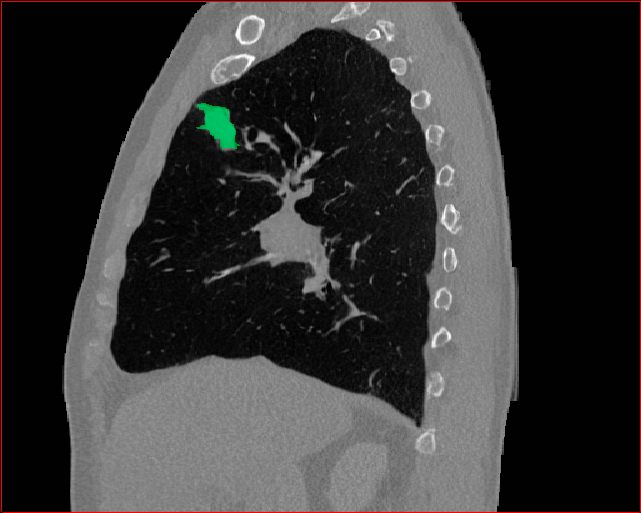}
\end{subfigure}
\begin{subfigure}[c]{0.195\textwidth}
    \includegraphics[width=\linewidth]{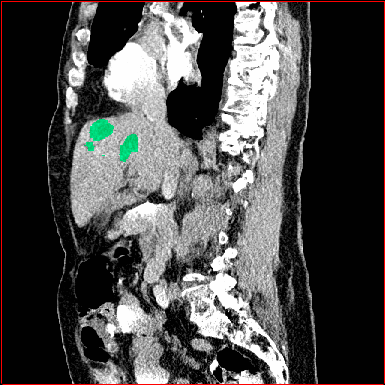}
\end{subfigure}
\begin{subfigure}[c]{0.195\textwidth}
    \includegraphics[width=\linewidth]{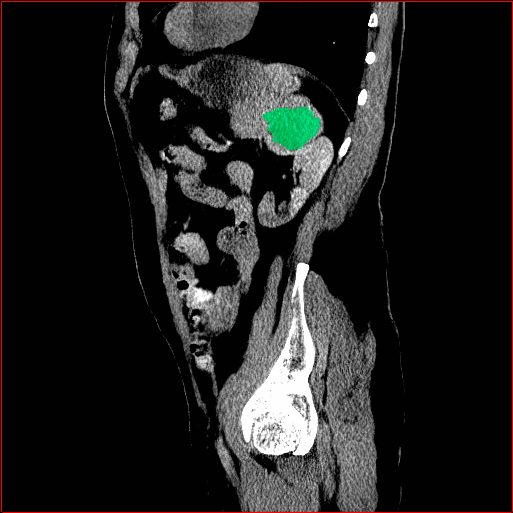}
\end{subfigure}
\begin{subfigure}[c]{0.195\textwidth}
    \includegraphics[width=\linewidth]{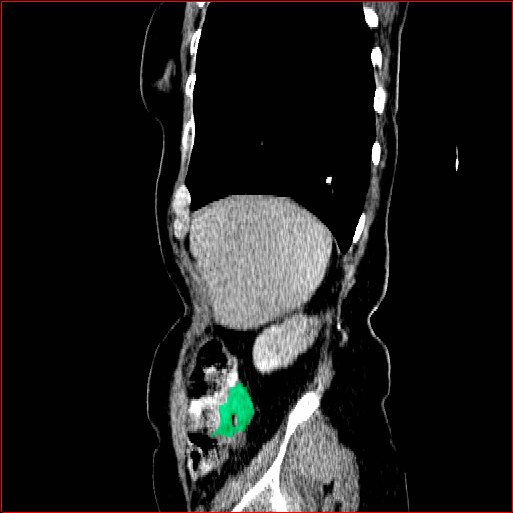}
\end{subfigure}
\begin{subfigure}[c]{0.195\textwidth}
    \includegraphics[width=\linewidth]{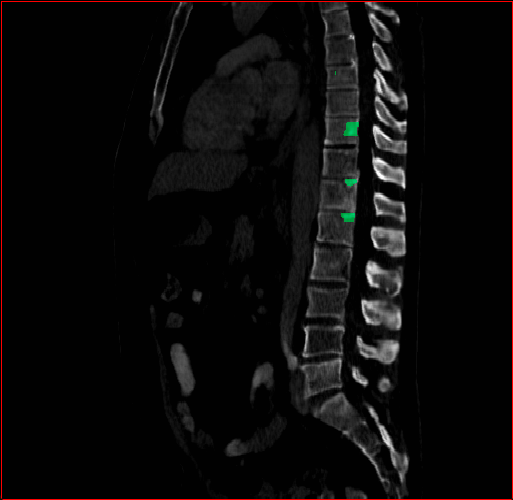}
\end{subfigure}

\begin{subfigure}[c]{0.195\textwidth}
    \includegraphics[width=\linewidth]{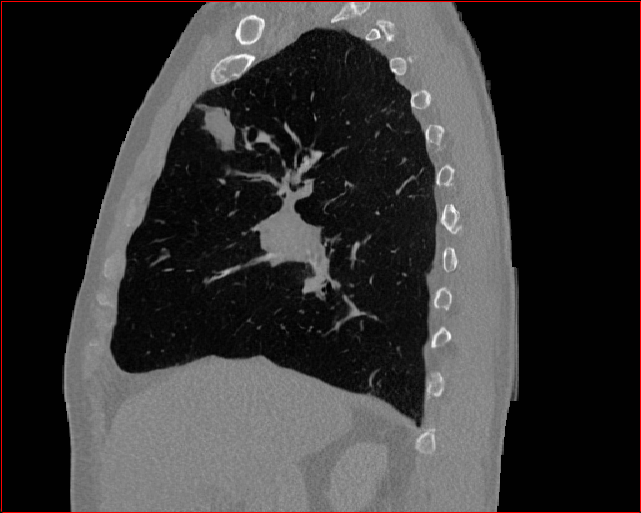}
\end{subfigure}
\begin{subfigure}[c]{0.195\textwidth}
    \includegraphics[width=\linewidth]{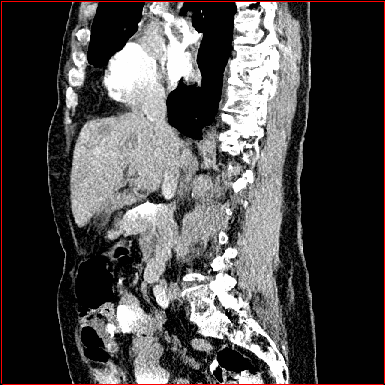}
\end{subfigure}
\begin{subfigure}[c]{0.195\textwidth}
    \includegraphics[width=\linewidth]{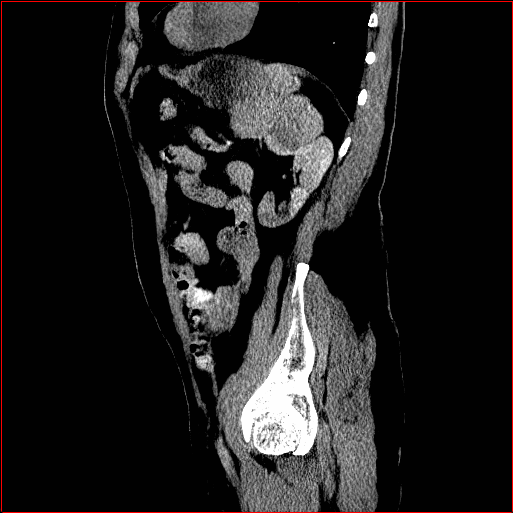}
\end{subfigure}
\begin{subfigure}[c]{0.195\textwidth}
    \includegraphics[width=\linewidth]{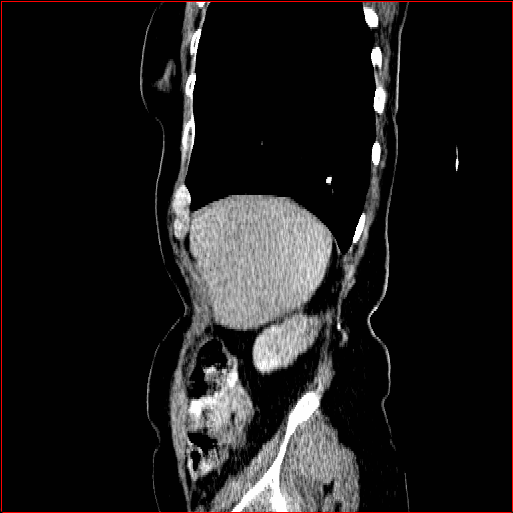}
\end{subfigure}
\begin{subfigure}[c]{0.195\textwidth}
    \includegraphics[width=\linewidth]{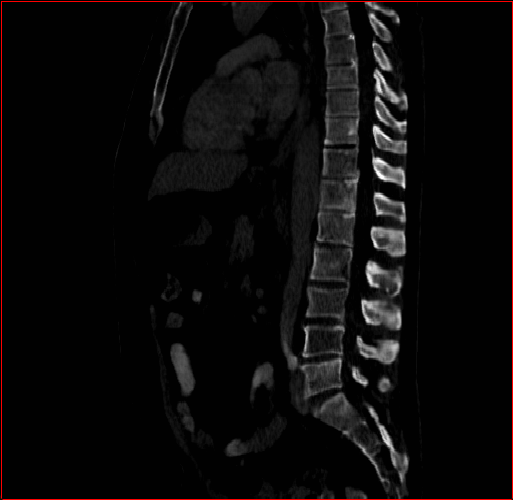}
\end{subfigure}

% Row 3: Coronal
\begin{subfigure}[c]{0.195\textwidth}
    \includegraphics[width=\linewidth]{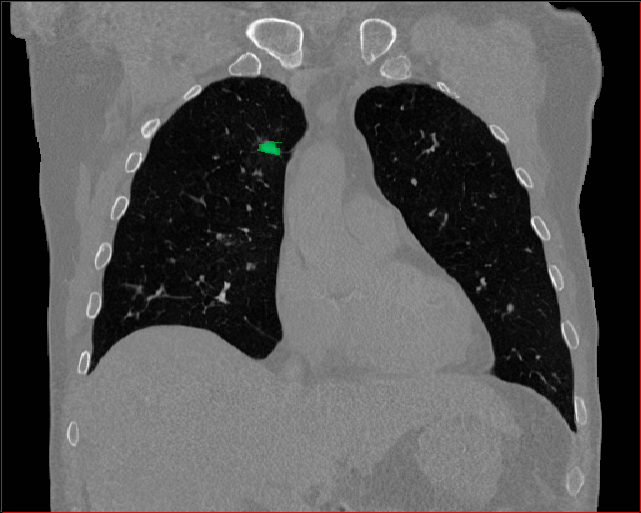}
\end{subfigure}
\begin{subfigure}[c]{0.195\textwidth}
    \includegraphics[width=\linewidth]{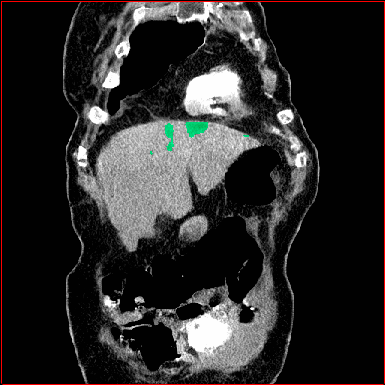}
\end{subfigure}
\begin{subfigure}[c]{0.195\textwidth}
    \includegraphics[width=\linewidth]{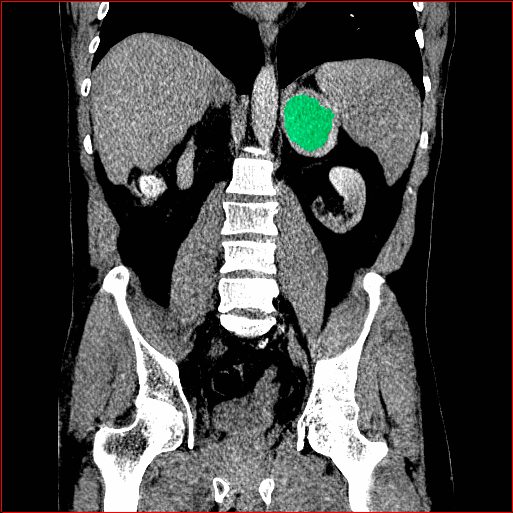}
\end{subfigure}
\begin{subfigure}[c]{0.195\textwidth}
    \includegraphics[width=\linewidth]{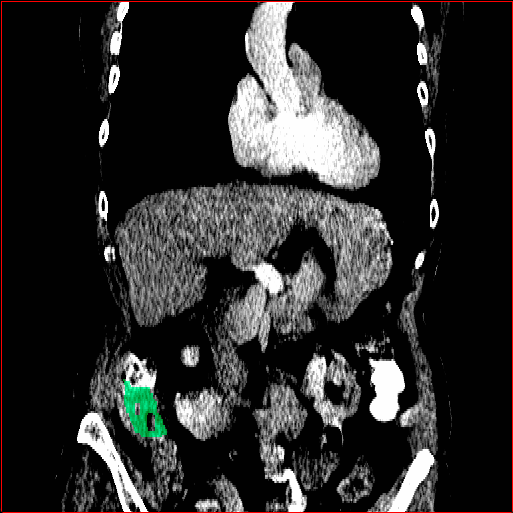}
\end{subfigure}
\begin{subfigure}[c]{0.195\textwidth}
    \includegraphics[width=\linewidth]{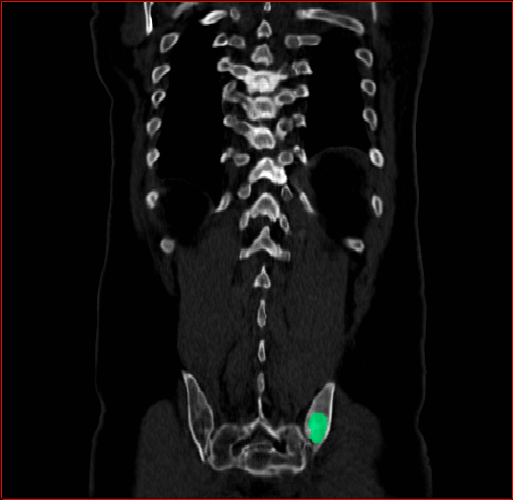}
\end{subfigure}

\begin{subfigure}[c]{0.195\textwidth}
    \includegraphics[width=\linewidth]{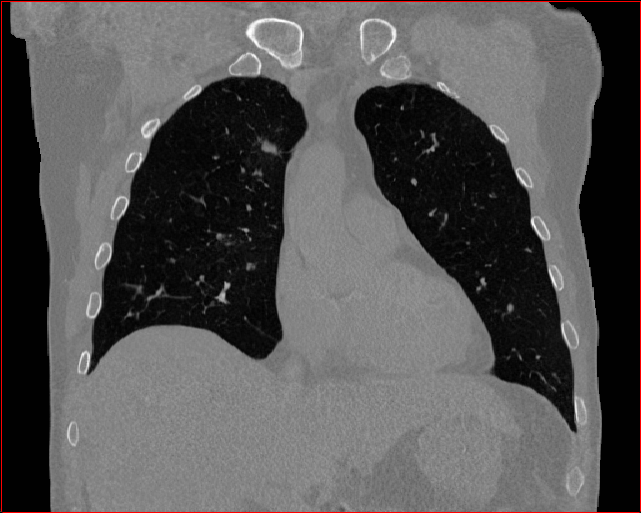}
\end{subfigure}
\begin{subfigure}[c]{0.195\textwidth}
    \includegraphics[width=\linewidth]{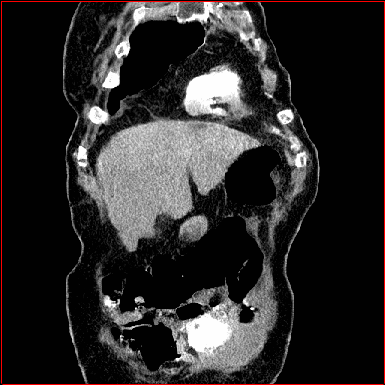}
\end{subfigure}
\begin{subfigure}[c]{0.195\textwidth}
    \includegraphics[width=\linewidth]{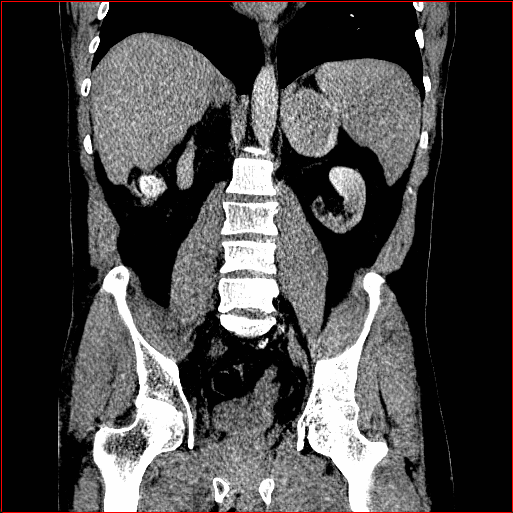}
\end{subfigure}
\begin{subfigure}[c]{0.195\textwidth}
    \includegraphics[width=\linewidth]{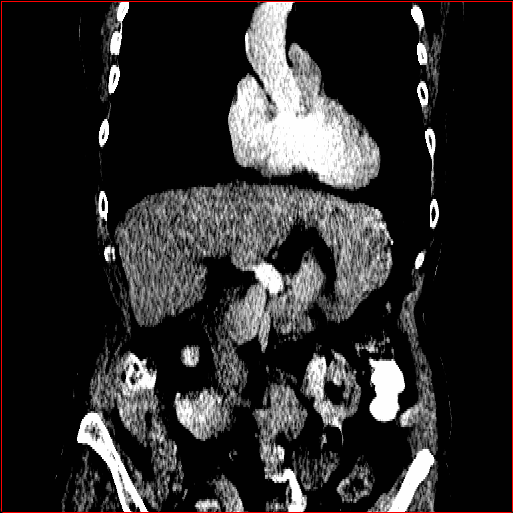}
\end{subfigure}
\begin{subfigure}[c]{0.195\textwidth}
    \includegraphics[width=\linewidth]{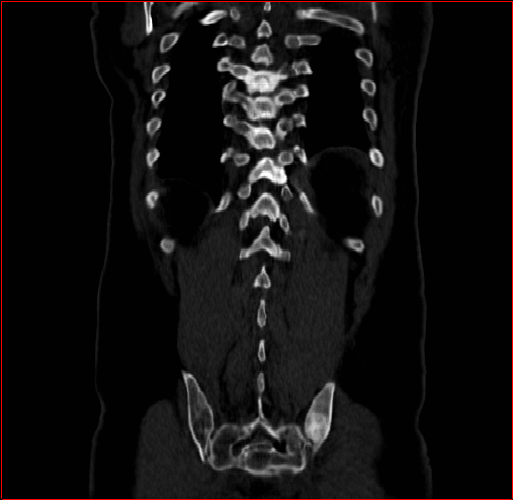}
\end{subfigure}
\caption{MAISI-v2 segmentation-guided results for five types of tumors. We show results for different voxel spacing and volume size to demonstrate the flexibility of MAISI-v2. Different Hounsfield Unit window is used to better show the contrast between tumor and normal tissues.}
\label{fig:tumor}
\end{figure*}

\paragraph{Qualitative Results:} Figure~\ref{fig:size} shows the generalizability of MAISI-v2 ControlNet for different body regions and voxel sizes. Figure~\ref{fig:tumor} shows qualitative results for MAISI-v2 ControlNet on 5 types of tumors.

%%%%%%%%%%%%%%%%%%%%%%%%%%%%

\begin{table*}[!htbp]
\centering
\caption{Segmentation Dice scores for five tumor tasks on the test sets. The baseline uses real images only. Others use real + synthetic images generated with different ControlNet training losses. ``(X\%)'' indicates improvement over the ``Real Only'' baseline. ``Panc. Tumor'' refers to Pancreatic Tumor.}
\label{tab:controlnet_segmentation}
\begin{tabular}{@{}lccccc@{}}
\toprule
\textbf{Method} & \textbf{Liver Tumor} & \textbf{Lung Tumor} & \textbf{Panc. Tumor} & \textbf{Colon Tumor} & \textbf{Bone-Les} \\
\midrule
Real Only & 0.662 & 0.581 & 0.433 & 0.449 & 0.504 \\
\midrule
DiffTumor & 0.684 (+2.2\%) & -- (--\%) & 0.511 (+7.9\%) & -- (--\%) & -- (--\%) \\
\midrule
MAISI     & 0.688 (+2.6\%) & 0.635 (+5.5\%) & 0.482 (+4.9\%) & 0.485 (+3.6\%) & 0.539 (+3.6\%) \\
\midrule
\multicolumn{6}{c}{\makecell[l]{\textbf{Ablation study:} ``MAISI-v2'' refers to ControlNet + Contra. loss. $\dagger$ means $p<0.1$, $*$ means $p<0.001$}} \\
ControlNet & 0.693 (+3.0\%) & 0.627 (+4.7\%) & 0.484 (+5.1\%) & 0.402 (-4.7\%) & 0.520 (+1.6\%) \\
MAISI-v2 & 0.695 (+3.3\%) & 0.655 (+7.5\%)$\dagger$ & 0.497 (+6.4\%)$\dagger$ & 0.491 (+4.2\%)* & 0.537 (+3.3\%)* \\
\bottomrule
\end{tabular}
\end{table*}

\paragraph{Quantitative Evaluation:}
To assess whether MAISI-generated images can improve downstream tasks, we perform a 5-fold segmentation data augmentation study.

We trained 3D segmentation networks with open-source tool Auto3DSeg~\footnote{\url{https://monai.io/apps/auto3dseg}. All the Auto3DSeg model training in this section share the same hyper-parameters.  } on: \textbf{(1) Baseline:} Real training images only. \textbf{(2) Augmented:} Real training images + same number of segmentation mask-guided synthetic images using different models, including DiffTumor~\cite{chen2024towards}, MAISI, and the proposed MAISI-v2. 

During inference, the input masks used for synthesis are not original tumor masks. Instead, we use those with tumors augmented with erosion, dilation and translation. Note that both MAISI and MAISI-v2 are single models that can generate five types of tumors, rather than five models for five types of tumors. DiffTumor~\cite{chen2024towards} is a diffusion model which was optimized specifically for tumor inpainting on the abdomen region for CT images. It takes healthy images and tumor masks as input and outputs images with inpainted tumors, while MAISI and MAISI-v2 only take segmentation masks as the input.

We compare the Dice scores on a held-out test set, reported on four public datasets—hepatic tumor (Task03)~\cite{antonelli2022medical}, lung tumor (Task06)~\cite{antonelli2022medical}, pancreatic tumor (Task07)~\cite{antonelli2022medical}, and colon tumor (Task10)~\cite{antonelli2022medical}, as well as one in-house dataset for bone lesions (Bone-Les), shown in Table~\ref{tab:controlnet_segmentation}. 

Results indicate that synthetic data generated by MAISI-v2 improves segmentation accuracy across multiple tumors, demonstrating its potential as a practical data augmentation tool in low-data scenarios.

\paragraph{Ablation Study:} We conduct an ablation study on the effect of the two proposed losses.
As shown in Table~\ref{tab:controlnet_segmentation}, it includes \textbf{(1) ControlNet}, with weighted average loss on tumors ($w=100$) and \textbf{(2) ControlNet with contrastive loss}, which is the proposed MAISI-v2. The final column reports the average improvement in Dice score over the real-only baseline. We performed a paired t-test on the values from the two methods. The improvement is significant for 4 out of 5 types of tumors.

\section{Conclusion and Discussion}
\label{s:conclusion}
\paragraph{Tumor alignment for Rectified Flow and DDPM:} We found that Rectified Flow ControlNet produces weaker tumor alignment than DDPM ControlNet during inference, especially for colon tumor. Two factors may contribute to this effect. First, Rectified Flow’s deterministic and straight-line transport formulation tends to reduce trajectory diversity~\cite{ma2025flow}, which may make the model less robust to small conditioning perturbations. Second, because it learns a single deterministic velocity field, prediction errors in small or low-contrast regions can accumulate and amplify during integration. In contrast, DDPM introduces stochastic noise at each reverse step, which helps diffuse such local errors and maintain robustness. Our region-specific contrastive loss mitigates both issues by enforcing local feature consistency and encouraging diverse mappings in the tumor regions. Future work will investigate the underlying mechanisms behind these observations.

\paragraph{Limitations and Future Work:} This work has three main limitations. (1) We only trained on CT scans. In the future, we plan to include other types like MRI and PET. (2) We only tested the model on segmentation tasks. Future work will include other applications like image translation, inpainting, detection, etc. (3) Few existing methods support high-resolution 3D synthesis due to GPU memory limitation. Although our method is already designed and optimized to train and infer on 24G GPU for small images like $512 \times 512 \times 256$, it needs large GPUs (40G for inference, 80G for training) for large images like $512 \times 512 \times 768$. In future work, we will further reduce memory use.

\paragraph{Broader Impacts:} We share our code, pretrained models, and training details, to make it easier for others to use or fine-tune. Like many other open-sourced large models, we hope to lower the barrier and reduce computing cost for future research in this community. 

\paragraph{Summary:} MAISI-v2 is an efficient and flexible method for 3D medical image synthesis. It uses rectified flow and new training losses to improve speed and condition accuracy. Our results show it creates high-quality images and helps improve segmentation models.

%%%%%%%%%%%%%%%%%%%%%%%%%%%%%%%%%%%%%%%%%%%%%%%%%%%%%%%%%%%%

\newpage
\section{Acknowledgments} 
We thank Weili Nie and Arash Vahdat for valuable discussions. The synthetic images from our model are not meant for clinical diagnosis. They are only used to help train other models, such as for segmentation, as shown in this paper.

\bibliography{aaai2026_final}

\end{document}